\pdfoutput=1

\documentclass[11pt]{article}

\usepackage[preprint]{acl}

\usepackage{times}
\usepackage{latexsym}

\usepackage[T1]{fontenc}

\usepackage[utf8]{inputenc}

\usepackage{microtype}

\usepackage{inconsolata}

\usepackage{graphicx}

\title{WangLab at MEDIQA-CORR 2024: Optimized LLM-based Programs for Medical Error Detection and Correction}

\author{
    Augustin Toma\textsuperscript{1,2} 
    Ronald Xie\textsuperscript{1,2} 
    Steven Palayew\textsuperscript{1,2} 
    Patrick R. Lawler\textsuperscript{1,3} 
    Bo Wang\textsuperscript{1,2,4,5} \\[5pt]
    \textsuperscript{1}University of Toronto \quad \textsuperscript{2}Vector Institute \\
    \textsuperscript{3}McGill University \quad \textsuperscript{4}Peter Munk Cardiac Centre, University Health Network \\
    \textsuperscript{5}AI Hub, University Health Network
    \\[5pt]
    \texttt{\{augustin.toma, ronald.xie, steven.palayew\}@mail.utoronto.ca}
    \\
    \texttt{bowang@vectorinstitute.ai}
}

\begin{document}
\maketitle 

\begin{abstract}
Medical errors in clinical text pose significant risks to patient safety. The MEDIQA-CORR 2024 shared task focuses on detecting and correcting these errors across three subtasks: identifying the presence of an error, extracting the erroneous sentence, and generating a corrected sentence. In this paper, we present our approach that achieved top performance in all three subtasks. For the MS dataset, which contains subtle errors, we developed a retrieval-based system leveraging external medical question-answering datasets. For the UW dataset, reflecting more realistic clinical notes, we created a pipeline of modules to detect, localize, and correct errors. Both approaches utilized the DSPy framework for optimizing prompts and few-shot examples in large language model (LLM) based programs. Our results demonstrate the effectiveness of LLM based programs for medical error correction. However, our approach has limitations in addressing the full diversity of potential errors in medical documentation. We discuss the implications of our work and highlight future research directions to advance the robustness and applicability of medical error detection and correction systems.
\end{abstract}
\section{Introduction}
Medical errors pose a significant threat to patient safety and can have severe consequences, including increased morbidity, mortality, and healthcare costs. Detecting and correcting these errors in clinical text is crucial for ensuring accurate medical documentation and facilitating effective communication among healthcare professionals. One of the fastest-growing use cases for artificial intelligence (AI) in healthcare is clinical note generation, often from transcriptions of physician-patient dialogues. However, assessing the quality and accuracy of these notes is challenging, and automated detection and correction of errors could have a significant impact on patient care.
The reliability of large language models (LLMs) in critical applications, such as healthcare, is a major concern due to the potential for hallucinations (generating false or nonsensical information) and inconsistencies. Robust solutions to the question of error detection and correction are essential for addressing these concerns and enabling the safe and effective use of LLMs in medical contexts.

The MEDIQA-CORR 2024 \citep{mediqa-corr-task} shared task focuses on identifying and correcting medical errors in clinical notes. Each text is either correct or contains a single error. The task involves three subtasks: (1) detecting the presence of an error, (2) extracting the erroneous sentence, and (3) generating a corrected sentence for flagged texts.

In this paper, we present our approach, which achieved the top performance across all three subtasks in the MEDIQA-CORR 2024 competition. We develop a series of LLM-based programs using DSPy, a framework for optimizing prompts and few-shot examples. We provide a detailed description of our methodology and results, followed by a discussion of the implications of our work and future directions in the field of medical error detection and correction.

\section{Related Work}
The use of large language models (LLMs) in medicine has attracted considerable attention in recent years. The release of LLMs such as GPT-4 has led to intensive research in the medical community \citep{nori2023capabilities}, particularly in clinical note generation. The MEDIQA-Chat 2023 \citep{ben-abacha-etal-2023-overview} competition showcased the performance of automated note generation solutions \citep{giorgi-etal-2023-wanglab}, and further work has demonstrated that LLMs can sometimes outperform humans on clinical text summarization tasks \citep{Van_Veen_2024}.

However, there has been limited research focusing on granular audits of these clinical notes with respect to accuracy and error correction. The MEDIQA-CORR 2024 shared task addresses this gap by providing a platform for researchers to develop and evaluate novel approaches to error detection and correction in clinical text, ultimately contributing to the development of more reliable AI systems in healthcare.

\section{Task Description}
The MEDIQA-CORR 2024 shared task provides two distinct datasets: MS and UW \citep{mediqa-corr-dataset}. The MS dataset consists of a Training Set containing 2,189 clinical texts and a Validation Set (\#1) containing 574 clinical texts. The UW dataset, on the other hand, consists solely of a Validation Set (\#2) containing 160 clinical texts. The test set for the shared task includes clinical texts from both the MS and UW collections.

The evaluation metrics for the MEDIQA-CORR 2024 shared task vary across the three subtasks:
\begin{itemize} \item Subtask 1 (Error Flag Prediction): Evaluated using Accuracy. \item Subtask 2 (Error Sentence Detection): Evaluated using Accuracy. \item Subtask 3 (Sentence Correction): Evaluated using ROUGE \cite{lin-2004-rouge}, BERTScore \cite{zhang2020bertscore}, BLEURT \cite{sellam2020bleurt}, Aggregate-Score (mean of ROUGE-1-F, BERTScore, BLEURT-20), and Composite Scores. \end{itemize}

The Composite Score for each text in Subtask 3 is calculated as follows:
\begin{enumerate} \item Assign 1 point if both the system correction and the reference correction are "NA" \item Assign 0 points if only one of the system correction or the reference correction is "NA" \item Calculate the score based on metrics (ROUGE, BERTScore, BLEURT and the Aggregate-Score) within the range of [0, 1] if both the system correction and reference correction are non-"NA" sentences. \end{enumerate}

\section{Approach}
\subsection{Overview}
Upon reviewing the MS and UW datasets, it became apparent that these two datasets presented distinct challenges. The errors in the MS dataset were often extremely subtle, to the point that many errors did not actually seem like errors, and in fact, clinicians on our team often couldn't identify the presence of an error within the text. However, when reviewing corrected text from the training set, it became clear that corrections were often 'optimal' completions. For example, consider the following error and its correction:

\begin{quote}
\textbf{Error sentence:} After reviewing imaging, the causal pathogen was determined to be Haemophilus influenzae. \citep{mediqa-corr-dataset}

\textbf{Corrected sentence:} After reviewing imaging, the causal pathogen was determined to be Streptococcus pneumoniae. \citep{mediqa-corr-dataset}
\end{quote}

These types of errors are subtle and seem akin to multiple-choice questions, where often multiple answers could independently be seen as correct completions, but only in the context of one another would you deem one answer wrong. On the other hand, the UW dataset appeared to reflect realistic clinical notes, and the errors were more apparent. For example, consider the following error and its correction:

\begin{quote}
\textbf{Error sentence:} Hypokalemia - based on laboratory findings patient has hypervalinemia. \citep{mediqa-corr-dataset}

\textbf{Corrected sentence:} Hypokalemia - based on laboratory findings patient has hypokalemia. \citep{mediqa-corr-dataset}
\end{quote}

In this case, the error involves a nonsensical term (hypervalinemia, a rare metabolic condition) when the context makes it clear that the patient has hypokalemia (low potassium levels). These are errors that a clinician can identify from the text alone. 

The distinct characteristics of the MS and UW datasets prompted us to develop a two-pronged approach to the MEDIQA-CORR 2024 shared task. For the MS dataset, we employed a retrieval-based system to identify similar questions from external medical question-answering datasets and leverage the knowledge contained in these datasets to detect and correct errors. For the UW dataset, we created a series of modules to detect, localize, and correct errors in clinical text snippets. Both approaches were built on DSPy \citep{khattab2023dspy}, a novel framework for systematically optimizing prompts and few-shot examples in LLM based programs.

\subsection{Approach for MS Dataset}
Our approach to the MS dataset involves a multi-step process that leverages retrieval-based methods and the DSPy framework, as illustrated in Figures \ref{fig:ms_step1}, \ref{fig:ms_step2}, and \ref{fig:ms_step3}. In all of our experiments, we utilized GPT-4-0125-preview as the underlying large language model, using default generation parameters (temperature of 1.0, top\_p of 1) with the exception of a max tokens value of 4096.

\subsubsection{Retrieval of Similar Questions}
First, we employ a retrieval-based approach to identify similar questions from the MedQA dataset \citep{jin2020disease}. MedQA is a medical question-answering dataset that contains multiple-choice questions, each with a set of answer options and a correct answer. By leveraging the knowledge contained in this external dataset, we aim to detect and correct errors in the MS dataset. We use TF-IDF \citep{sparck1972statistical} to calculate the similarity between the given question in the MS dataset and the questions in MedQA, retrieving the most similar questions along with their answer options and correct answers for further analysis.

\subsubsection{Identifying Answer Choices within Query Text}
To identify the implicit answer choice within the query text, we employ a two-step process using DSPy programs. First, we send both the query text and the identified similar multiple-choice question to a DSPy module that utilizes chain of thought \citep{wei2023chainofthought} and the BootstrapFewShotWithRandomSearch teleprompter \citep{khattab2023dspy}. This teleprompter generates 20 few-shot examples by sampling from the training set and testing the module's performance on the validation set. The module aims to extract the answer choice that appears to be present in the query text.

\begin{figure}[t]
  \includegraphics[width=\columnwidth]{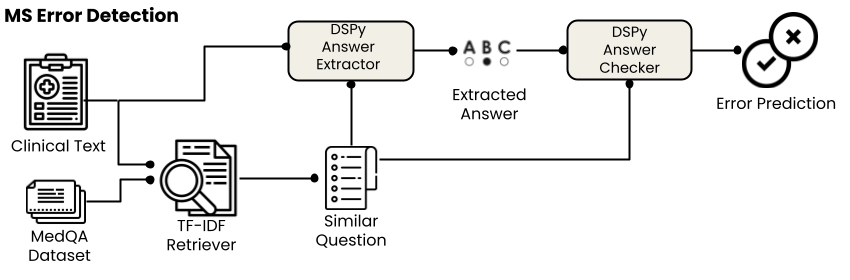}
  \caption{Predicting the presence of an error through a comparison to the retrieved question}
  \label{fig:ms_step1}
\end{figure}

The output from this module is then passed to a second DSPy module, which also leverages the BootstrapFewShotWithRandomSearch teleprompter. This module creates multiple few-shot examples that compare the extracted answer against the true answer from the multiple-choice question, as shown in Figure \ref{fig:ms_step1}. We simultaneously bootstrap these two steps, optimizing the entire pipeline based on the accuracy of the overall error flag prediction.

The result of this bootstrapping process is a compiled program with optimized multi-step chain of thought prompts based on the module's performance on error detection accuracy. This approach allows us to effectively identify the presence of errors in the query text by leveraging the knowledge from external medical question-answering datasets.

\subsubsection{Localizing Errors within Query Text}
\begin{figure}[t]
  \includegraphics[width=\columnwidth]{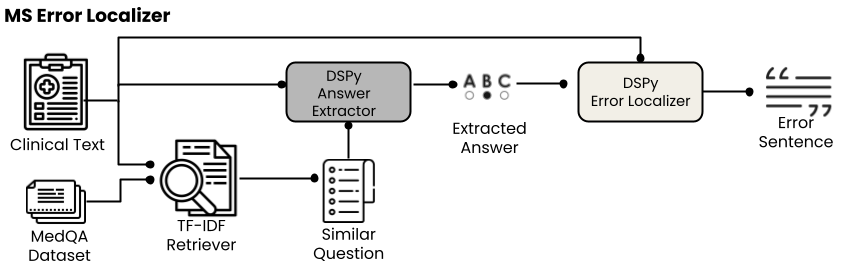}
  \caption{Identifying the error sentence}
  \label{fig:ms_step2}
\end{figure}

After detecting an error in the query text, we use a DSPy module to identify the specific line containing the error, as illustrated in Figure \ref{fig:ms_step2}. This module takes the extracted answer choice and the preprocessed query text as inputs and then an LLM call is done to determine which line most closely matches the erroneous answer choice.

Our experiments showed that GPT-4's performance was high enough that we did not need to compile the program or bootstrap few-shot prompts via a DSPy teleprompter.

The module outputs the line number where the error is located, which is crucial for the subsequent error correction step, as it allows for targeted correction of the relevant text.

\subsubsection{Error Correction with DSPy}
\begin{figure}[t]
  \includegraphics[width=\columnwidth]{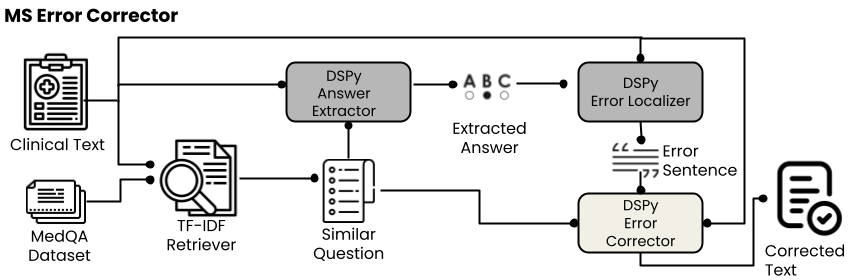}
  \caption{Generating the corrected sentence}
  \label{fig:ms_step3}
\end{figure}

After identifying the error location within the query text, we use a final DSPy module to generate a corrected version of the text, as illustrated in Figure \ref{fig:ms_step3}. This module takes three inputs: the error line, the extracted answer choice, and the correct answer derived from the most similar retrieved multiple-choice question.

The error correction module utilizes a chain of thought prompt along with 20 few-shot examples generated by the BootstrapFewShotWithRandomSearch teleprompter. This teleprompter samples examples from the training set and generates intermediate labels, such as rationales for the chain of thought, to provide additional context and guidance for the language model during the error correction process. The teleprompter optimizes the selection of few-shot prompts based on their performance on the validation set, using the ROUGE-L score as the metric.

The selected few-shot examples, accompanied by the generated intermediate labels, demonstrate how to modify the error line based on the extracted answer choice and the correct answer, serving as a reference for the model to learn from and adapt to the specific error correction task.

The module outputs the corrected version of the query text, with the error line revised based on the correct answer derived from the most similar multiple-choice question. This corrected text represents the final output of our retrieval-based approach for the MS dataset, addressing the subtle errors present in the clinical text.

\subsection{Approach for UW Dataset}
\begin{figure}[t]
  \includegraphics[width=\columnwidth]{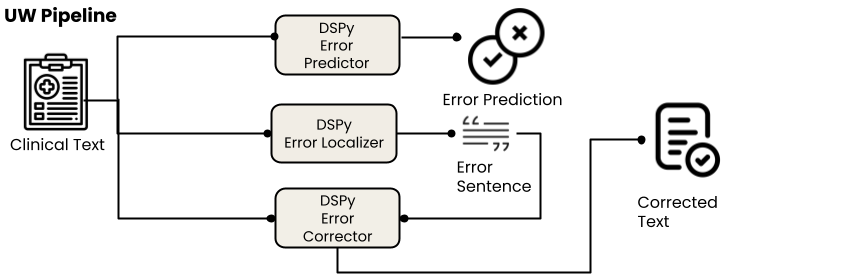}
  \caption{Overview of the UW dataset pipeline, consisting of three main stages: error detection, error localization, and error correction. Each stage is implemented using a DSPy module optimized with the MIPRO teleprompter \citep{khattab2023dspy} The pipeline also includes a quality control step based on the ROUGE-L score between the original erroneous text and the corrected version.}
  \label{fig:uw_pipeline}
\end{figure}

Our approach for the UW dataset involves optimizing a series of DSPy modules to accomplish all three subtasks sequentially, as illustrated in Figure \ref{fig:uw_pipeline}. In all of our experiments, we utilized GPT-4-0125-preview as the underlying large language model, using default generation parameters (temperature of 1.0, top\_p of 1) with the exception of a max tokens value of 4096.

\subsubsection{Error Detection with DSPy}
For the UW dataset, we first employ a DSPy program to identify whether an error exists in the given clinical text snippet. This program is optimized using the Multi-prompt Instruction Proposal Optimizer (MIPRO) teleprompter, which generates and optimizes both the base prompts and few-shot examples. MIPRO optimizes the prompts and few-shot examples to maximize performance on the validation set, which we created by dividing the UW training collection (160 examples) into 80 training examples, 40 validation examples, and 40 test examples. The optimizer uses error flag accuracy as the metric to optimize and generates 20 examples. We also incorporate chain of thought reasoning into the DSPy module.

\subsubsection{Error Localization}
If an error is detected in the clinical text snippet, we use another DSPy module to identify the specific line containing the error. This module is also optimized using MIPRO, which generates 20 bootstrap examples that include chain of thought rationales. Using a separate DSPy module for error localization allows us to precisely identify the source of the error and facilitate targeted corrections. The exact match of the error line is used as the metric for optimization, and this module is trained only on a subset of the training samples that contain errors.

\subsubsection{Error Correction}
After identifying the error line, we use a third DSPy module to generate a corrected version of the erroneous text. This module is also optimized using MIPRO, following the same process as the previous modules. The error correction module takes the erroneous text as input and generates a corrected version based on the optimized prompts and weights. MIPRO uses the ROUGE-L score against the known correct sentence as the metric to optimize, and this module is trained only on a subset of the training samples that contain errors.

\subsubsection{Quality Control with ROUGE-L}
To ensure the quality of the generated corrections, we calculate the ROUGE-L score between the original erroneous text and the corrected version. If the ROUGE-L score is below a threshold of 0.7, which we set as an arbitrary estimate for quality, we reject the correction and use the original erroneous text instead. This fallback mechanism is based on the observation that the ROUGE-L score of the erroneous text tends to be quite high since the error is only a small portion of the sentence. However, this fallback is more of a contest-metric-focused feature rather than something that significantly improves performance.

\section{Results and Discussion}
\subsection{Overall Performance in the MEDIQA-CORR 2024 Shared Task}
Our approach achieved top performance in the MEDIQA-CORR 2024 shared task across all three subtasks. Tables \ref{tab:task1_top10}, \ref{tab:task2_top10}, and \ref{tab:task3_top10} present the performance of the top 10 teams in each subtask.

\subsection{Performance on Subtask 1 - Error Prediction}
In the official contest results for binary error prediction, our approach achieved an accuracy of 86.5\%, ranking first among all participating teams. Table \ref{tab:task1_top10} shows the top 10 teams' performance on Task 1.

\begin{table}[t]
\centering
\resizebox{\columnwidth}{!}{%
\begin{tabular}{clc}
\hline
Rank & Team & Error Flags Accuracy \\ \hline
1 & WangLab & 86.5\% \\
2 & MediFact & 73.7\% \\
3 & knowlab\_AIMed & 69.4\% \\
4 & EM\_Mixers & 68.0\% \\
5 & IKIM & 67.8\% \\
6 & IryoNLP & 67.1\% \\
7 & Edinburgh Clinical NLP & 66.9\% \\
8 & hyeonhwang & 63.5\% \\
9 & PromptMind & 62.2\% \\
10 & CLD-MEC & 56.6\% \\
\hline
\end{tabular}%
}
\caption{Top 10 teams' performance on Task 1 (Error Flags Accuracy)}
\label{tab:task1_top10}
\end{table}

\subsection{Performance on Subtask 2 - Error Sentence Detection}
For error sentence detection, we obtained an accuracy of  83.6\%, ranking first among all teams. Table \ref{tab:task2_top10} presents the top 10 teams' performance.

\begin{table}[t]
\centering
\resizebox{\columnwidth}{!}{%
\begin{tabular}{clc}
\hline
Rank & Team & Error Sentence Detection Accuracy \\ \hline
1 & WangLab & 83.6\% \\
2 & EM\_Mixers & 64.0\% \\
3 & knowlab\_AIMed & 61.9\% \\
4 & hyeonhwang & 61.5\% \\
5 & Edinburgh Clinical NLP & 61.1\% \\
6 & IryoNLP & 61.0\% \\
7 & PromptMind & 60.9\% \\
8 & MediFact & 60.0\% \\
9 & IKIM & 59.0\% \\
10 & HSE NLP & 52.0\% \\
\hline
\end{tabular}%
}
\caption{Top 10 teams' performance on Task 2 (Error Sentence Detection Accuracy)}
\label{tab:task2_top10}
\end{table}

These results demonstrate the effectiveness of our few-shot learning and CoT-based approach in detecting the presence of errors and localizing the specific sentences containing the errors.

\subsection{Performance on Subtask 3 - Sentence Correction}
For subtask C (Sentence Correction), the official contest results show that our approach achieved an Aggregate-Score of 0.789, which is the mean of ROUGE-1-F (0.776), BERTScore (0.809), and BLEURT (0.783). This was the highest score among the participating teams for the sentence correction task. Table \ref{tab:task3_top10} displays the top 10 teams' performance on Task 3.

\begin{table*}[t]
\centering
\resizebox{\textwidth}{!}{%
\begin{tabular}{clccccc}
\hline
Rank & Team & AggregateScore & R1F & BERTSCORE & BLEURT & AggregateCR \\ \hline
1 & WangLab & 0.789 & 0.776 & 0.809 & 0.783 & 0.775 \\
2 & PromptMind & 0.787 & 0.807 & 0.806 & 0.747 & 0.574 \\
3 & HSE NLP & 0.781 & 0.779 & 0.806 & 0.756 & 0.512 \\
4 & hyeonhwang & 0.734 & 0.729 & 0.767 & 0.705 & 0.571 \\
5 & Maven & 0.733 & 0.703 & 0.744 & 0.752 & 0.524 \\
6 & Edinburgh Clinical NLP & 0.711 & 0.678 & 0.744 & 0.711 & 0.563 \\
7 & knowlab\_AIMed & 0.658 & 0.643 & 0.677 & 0.654 & 0.573 \\
8 & EM\_Mixers & 0.587 & 0.571 & 0.595 & 0.596 & 0.548 \\
9 & IryoNLP & 0.581 & 0.561 & 0.592 & 0.591 & 0.528 \\
10 & IKIM & 0.559 & 0.523 & 0.564 & 0.588 & 0.550 \\
\hline
\end{tabular}%
}
\caption{Top 10 teams' performance on Task 3 (Aggregate Score and its components)}
\label{tab:task3_top10}
\end{table*}

The official contest results highlight the competitive performance of our approach across all three subtasks of the MEDIQA-CORR 2024 shared task, demonstrating its effectiveness in detecting, localizing, and correcting medical errors in clinical text for both the MS and UW datasets.

\subsection{Implications and Limitations of the Approach}
Our work contributes to the ongoing efforts in improving the accuracy and reliability of medical information in clinical text. The automated detection and correction of certain types of errors could ensure the quality and consistency of medical documentation, ultimately supporting patient safety and quality of care. The development and integration of more advanced systems could help alleviate the burden of manual error checking for the specific error types addressed, allowing healthcare providers to allocate more time and resources to delivering high-quality patient care.

However, it is important to acknowledge the limitations of our approach in the context of the diverse nature of errors in medical documentation. While our system demonstrates strong performance on the MS and UW datasets, it focuses on a specific subset of errors and has not been shown to be effective in addressing the wide diversity of errors that can occur in medical documentation.

For instance, our approach does not currently address errors that are propagated through multiple notes when a physician references prior documents containing inaccuracies, such as incorrect medical history. Such errors can be particularly challenging to identify and correct, as they may require a comprehensive understanding of the patient's medical history, the context of the referenced documents, and the resolution of conflicting statements across documents. Our system has not been designed or evaluated for handling these types of errors.

Moreover, our approach does not cover errors that originate from sources beyond the scope of our training data, such as poor transcriptions, entries in the wrong medical record, or errors in decision making. These types of errors may necessitate different strategies and techniques for detection and correction, and our current approach has not been developed to handle them.

Additionally, the reliance on external datasets for the retrieval-based approach in the MS dataset limits the generalizability of our method to other medical domains or datasets. In fact, we believe that an approach used in the MS dataset might actually create further errors if used on real clinical text, as real clinical practice does not always reflect optimal or most likely completions. The effectiveness of our approach in detecting and correcting errors may vary depending on the specific characteristics and error types present in different medical contexts, and further evaluation would be necessary to assess its performance in diverse settings.

\begin{table*}[t]
\centering
\resizebox{\textwidth}{!}{%
\begin{tabular}{lcccc}
\hline
\multicolumn{5}{c}{\textbf{Error Flags Accuracy (Task 1)}} \\
\hline
 & GPT-3.5 Compiled & GPT-3.5 Uncompiled & GPT-4 Compiled & GPT-4 Uncompiled \\
\hline
Error Flags Accuracy & 94.0\% (0.4\%) & 81.2\% (0.7\%) & 97.3\% (0.1\%) & 88.9\% (0.5\%) \\
\hline
\multicolumn{5}{c}{\textbf{Error Sentence Detection Accuracy (Task 2)}} \\
\hline
 & GPT-3.5 Compiled & GPT-3.5 Uncompiled & GPT-4 Compiled & GPT-4 Uncompiled \\
\hline
Error Sentence Detection Accuracy & 92.8\% (0.5\%) & 78.5\% (0.8\%) & 97.0\% (0.1\%) & 88.0\% (0.8\%) \\
\hline
\multicolumn{5}{c}{\textbf{Task 3 Metrics}} \\
\hline
Metric & GPT-3.5 Compiled & GPT-3.5 Uncompiled & GPT-4 Compiled & GPT-4 Uncompiled \\
\hline
aggregate\_subset\_check & 0.853 (0.001) & 0.809 (0.011) & 0.824 (0.003) & 0.827 (0.003) \\
R1F\_subset\_check & 0.827 (0.003) & 0.778 (0.017) & 0.789 (0.003) & 0.792 (0.003) \\
BERTSCORE\_subset\_check & 0.874 (0.001) & 0.827 (0.013) & 0.856 (0.003) & 0.857 (0.002) \\
BLEURT\_subset\_check & 0.859 (0.000) & 0.824 (0.006) & 0.827 (0.002) & 0.832 (0.003) \\
AggregateC & 0.864 (0.004) & 0.736 (0.010) & 0.878 (0.002) & 0.792 (0.005) \\
\hline
\end{tabular}%
}
\caption{Ablation studies for error flag accuracy (Task 1), error sentence detection accuracy (Task 2), and Task 3 metrics. Numbers in parentheses represent standard deviations.}
\label{tab:ablation_combined}
\end{table*}

\subsubsection{Impact of Different LLMs and Compilation}
After the competition ended, we performed additional experiments to compare the performance of our approach when using GPT-4 and GPT-3.5 as the underlying language models for the DSPy modules, as well as the impact of using compiled and uncompiled DSPy programs.

Table \ref{tab:ablation_combined} presents the results of the ablation study for error flag accuracy (Task 1), error sentence detection accuracy (Task 2), and various metrics for Task 3. The results show that using GPT-4 as the underlying LLM consistently yields better performance compared to GPT-3.5 across all tasks. For Task 1, the compiled GPT-4 model achieves the highest accuracy of 97.3\% (0.1\%), while for Task 2, it achieves an accuracy of 97.0\% (0.1\%). The compiled DSPy programs outperform their uncompiled counterparts for both GPT-3.5 and GPT-4.

In Task 3, the compiled GPT-4 model consistently outperforms the other models across all metrics, with the highest AggregateC score of 0.878 (0.002). Moreover, the results demonstrate that using compiled DSPy programs consistently outperforms the uncompiled approach across all tasks and datasets, emphasizing the significance of systematic optimization techniques in enhancing the performance of our error detection and correction system.

It is important to note that we did not isolate the impact of retrieval in our post-competition experiments, as it was a fundamental component of all the modules in our approach. Removing the retrieval component would require the development of a new solution. However, the strong performance of our uncompiled GPT-3.5 solution suggests that a significant portion of the performance could be attributed to the retrieval process itself. Future work should explore the impact of different retrieval strategies on the performance of error detection and correction in clinical text.

\subsection{Future Research Directions}
Although our approach has demonstrated competitive performance in the MEDIQA-CORR 2024 shared task, there are several potential avenues for future research that could further improve the effectiveness and applicability of our system.

One area for future investigation is the fine-tuning of open access models specifically for clinical notes \citep{toma2023clinical}. While fine-tuning may lead to higher performance, we focused on working with DSPy in the current study and did not have the computational resources to maintain the necessary throughput and latency during initial experimentation. Future studies could examine the trade-offs between fine-tuning and using off-the-shelf models with prompt optimization techniques, taking into account factors such as performance, efficiency, and scalability.

Another direction for future research is the expansion of the benchmark dataset to include a broader range of errors, such as those spanning multiple documents or involving suboptimal clinical decisions. Broadening the scope of the dataset would enhance the robustness of error detection and correction systems and extend their applicability to more complex clinical scenarios.

Integrating domain-specific knowledge, such as medical ontologies or expert-curated rules, into our approach could improve the system's ability to handle complex medical cases and make more informed decisions. This would be particularly relevant if the errors include suboptimal clinical decisions, as the system could provide more comprehensive support to healthcare professionals.

Lastly, developing more comprehensive and robust methods for measuring and correcting errors is an area with significant potential. This could involve creating standardized evaluation metrics and datasets that better capture the intricacies of medical errors and developing more advanced error correction techniques that can handle a wider range of error types and contexts.

\section{Conclusion}
The approach presented in this paper, which combines retrieval-based methods, few-shot learning, and systematic prompt optimization, demonstrates the potential of AI-assisted tools for detecting and correcting medical errors in clinical text. The strong performance achieved across all three subtasks of the MEDIQA-CORR 2024 shared task highlights the effectiveness of our methods in addressing the specific challenges posed by different datasets and error types. However, further research is necessary to extend the applicability of our approach to a wider range of medical contexts, incorporate domain-specific knowledge, and integrate with existing clinical systems. As the field of AI-assisted medical error detection and correction continues to evolve, collaboration between AI researchers and healthcare professionals will be crucial to develop solutions that effectively augment and support clinical decision-making processes, ultimately contributing to improved patient safety and healthcare quality.

\clearpage
\bibliography{custom}

\begin{thebibliography}{14}
\providecommand{\natexlab}[1]{#1}

\bibitem[{{Ben Abacha} et~al.(2024{\natexlab{a}}){Ben Abacha}, wai Yim, Fu, Sun, Xia, and Yetisgen}]{mediqa-corr-task}
Asma {Ben Abacha}, Wen wai Yim, Yujuan Fu, Zhaoyi Sun, Fei Xia, and Meliha Yetisgen. 2024{\natexlab{a}}.
\newblock Overview of the mediqa-corr 2024 shared task on medical error detection and correction.
\newblock In \emph{Proceedings of the 6th Clinical Natural Language Processing Workshop}, Mexico City, Mexico. Association for Computational Linguistics.

\bibitem[{{Ben Abacha} et~al.(2024{\natexlab{b}}){Ben Abacha}, wai Yim, Fu, Sun, Yetisgen, Xia, and Lin}]{mediqa-corr-dataset}
Asma {Ben Abacha}, Wen wai Yim, Yujuan Fu, Zhaoyi Sun, Meliha Yetisgen, Fei Xia, and Thomas Lin. 2024{\natexlab{b}}.
\newblock Medec: A benchmark for medical error detection and correction in clinical notes.
\newblock \emph{CoRR}.

\bibitem[{Ben~Abacha et~al.(2023)Ben~Abacha, Yim, Adams, Snider, and Yetisgen}]{ben-abacha-etal-2023-overview}
Asma Ben~Abacha, Wen-wai Yim, Griffin Adams, Neal Snider, and Meliha Yetisgen. 2023.
\newblock \href {https://doi.org/10.18653/v1/2023.clinicalnlp-1.52} {Overview of the {MEDIQA}-chat 2023 shared tasks on the summarization {\&} generation of doctor-patient conversations}.
\newblock In \emph{Proceedings of the 5th Clinical Natural Language Processing Workshop}, pages 503--513, Toronto, Canada. Association for Computational Linguistics.

\bibitem[{Giorgi et~al.(2023)Giorgi, Toma, Xie, Chen, An, Zheng, and Wang}]{giorgi-etal-2023-wanglab}
John Giorgi, Augustin Toma, Ronald Xie, Sondra Chen, Kevin An, Grace Zheng, and Bo~Wang. 2023.
\newblock \href {https://doi.org/10.18653/v1/2023.clinicalnlp-1.36} {{W}ang{L}ab at {MEDIQA}-chat 2023: Clinical note generation from doctor-patient conversations using large language models}.
\newblock In \emph{Proceedings of the 5th Clinical Natural Language Processing Workshop}, pages 323--334, Toronto, Canada. Association for Computational Linguistics.

\bibitem[{Jin et~al.(2020)Jin, Pan, Oufattole, Weng, Fang, and Szolovits}]{jin2020disease}
Di~Jin, Eileen Pan, Nassim Oufattole, Wei-Hung Weng, Hanyi Fang, and Peter Szolovits. 2020.
\newblock \href {https://arxiv.org/abs/2009.13081} {What disease does this patient have? a large-scale open domain question answering dataset from medical exams}.
\newblock \emph{Preprint}, arXiv:2009.13081.

\bibitem[{Khattab et~al.(2023)Khattab, Singhvi, Maheshwari, Zhang, Santhanam, Vardhamanan, Haq, Sharma, Joshi, Moazam, Miller, Zaharia, and Potts}]{khattab2023dspy}
Omar Khattab, Arnav Singhvi, Paridhi Maheshwari, Zhiyuan Zhang, Keshav Santhanam, Sri Vardhamanan, Saiful Haq, Ashutosh Sharma, Thomas~T. Joshi, Hanna Moazam, Heather Miller, Matei Zaharia, and Christopher Potts. 2023.
\newblock \href {https://arxiv.org/abs/2310.03714} {Dspy: Compiling declarative language model calls into self-improving pipelines}.
\newblock \emph{Preprint}, arXiv:2310.03714.

\bibitem[{Lin(2004)}]{lin-2004-rouge}
Chin-Yew Lin. 2004.
\newblock \href {https://aclanthology.org/W04-1013} {{ROUGE}: A package for automatic evaluation of summaries}.
\newblock In \emph{Text Summarization Branches Out}, pages 74--81, Barcelona, Spain. Association for Computational Linguistics.

\bibitem[{Nori et~al.(2023)Nori, King, McKinney, Carignan, and Horvitz}]{nori2023capabilities}
Harsha Nori, Nicholas King, Scott~Mayer McKinney, Dean Carignan, and Eric Horvitz. 2023.
\newblock \href {https://arxiv.org/abs/2303.13375} {Capabilities of gpt-4 on medical challenge problems}.
\newblock \emph{Preprint}, arXiv:2303.13375.

\bibitem[{Sellam et~al.(2020)Sellam, Das, and Parikh}]{sellam2020bleurt}
Thibault Sellam, Dipanjan Das, and Ankur~P. Parikh. 2020.
\newblock \href {https://arxiv.org/abs/2004.04696} {Bleurt: Learning robust metrics for text generation}.
\newblock \emph{Preprint}, arXiv:2004.04696.

\bibitem[{Sparck~Jones(1972)}]{sparck1972statistical}
Karen Sparck~Jones. 1972.
\newblock A statistical interpretation of term specificity and its application in retrieval.
\newblock \emph{Journal of documentation}, 28(1):11--21.

\bibitem[{Toma et~al.(2023)Toma, Lawler, Ba, Krishnan, Rubin, and Wang}]{toma2023clinical}
Augustin Toma, Patrick~R. Lawler, Jimmy Ba, Rahul~G. Krishnan, Barry~B. Rubin, and Bo~Wang. 2023.
\newblock \href {https://arxiv.org/abs/2305.12031} {Clinical camel: An open expert-level medical language model with dialogue-based knowledge encoding}.
\newblock \emph{Preprint}, arXiv:2305.12031.

\bibitem[{Van~Veen et~al.(2024)Van~Veen, Van~Uden, Blankemeier, Delbrouck, Aali, Bluethgen, Pareek, Polacin, Reis, Seehofnerová, Rohatgi, Hosamani, Collins, Ahuja, Langlotz, Hom, Gatidis, Pauly, and Chaudhari}]{Van_Veen_2024}
Dave Van~Veen, Cara Van~Uden, Louis Blankemeier, Jean-Benoit Delbrouck, Asad Aali, Christian Bluethgen, Anuj Pareek, Malgorzata Polacin, Eduardo~Pontes Reis, Anna Seehofnerová, Nidhi Rohatgi, Poonam Hosamani, William Collins, Neera Ahuja, Curtis~P. Langlotz, Jason Hom, Sergios Gatidis, John Pauly, and Akshay~S. Chaudhari. 2024.
\newblock \href {https://doi.org/10.1038/s41591-024-02855-5} {Adapted large language models can outperform medical experts in clinical text summarization}.
\newblock \emph{Nature Medicine}.

\bibitem[{Wei et~al.(2023)Wei, Wang, Schuurmans, Bosma, Ichter, Xia, Chi, Le, and Zhou}]{wei2023chainofthought}
Jason Wei, Xuezhi Wang, Dale Schuurmans, Maarten Bosma, Brian Ichter, Fei Xia, Ed~Chi, Quoc Le, and Denny Zhou. 2023.
\newblock \href {https://arxiv.org/abs/2201.11903} {Chain-of-thought prompting elicits reasoning in large language models}.
\newblock \emph{Preprint}, arXiv:2201.11903.

\bibitem[{Zhang et~al.(2020)Zhang, Kishore, Wu, Weinberger, and Artzi}]{zhang2020bertscore}
Tianyi Zhang, Varsha Kishore, Felix Wu, Kilian~Q. Weinberger, and Yoav Artzi. 2020.
\newblock \href {https://arxiv.org/abs/1904.09675} {Bertscore: Evaluating text generation with bert}.
\newblock \emph{Preprint}, arXiv:1904.09675.

\end{thebibliography}

\end{document}